\documentclass[runningheads]{llncs}
\usepackage{graphicx}
\begin{document}
\title{EPIE Dataset: A Corpus For Possible Idiomatic Expressions}
%
%
\author{Prateek Saxena\orcidID{0000-0001-9628-3858} \and
 Soma Paul\orcidID{0000-0002-2504-4419}}
\institute{International Institute of Information Technology, Hyderabad}

%
\maketitle              
\begin{abstract}
Idiomatic expressions have always been a bottleneck for language comprehension and natural language understanding, specifically for tasks like Machine Translation(MT). MT systems predominantly produce literal translations of idiomatic expressions as they do not exhibit generic and linguistically deterministic patterns which can be exploited for comprehension of the non-compositional meaning of the expressions. These expressions occur in parallel corpora used for training, but due to the comparatively high occurrences of the constituent words of idiomatic expressions in literal context, the idiomatic meaning gets overpowered by the compositional meaning of the expression. State of the art Metaphor Detection Systems are able to detect non-compositional usage at word level but miss out on idiosyncratic phrasal idiomatic expressions. This creates a dire need for a dataset with a wider coverage and higher occurrence of commonly occurring idiomatic expressions, the spans of which can be used for Metaphor Detection. With this in mind, we present our English Possible Idiomatic Expressions(EPIE) corpus containing 25206 sentences labelled with lexical instances of 717 idiomatic expressions. These spans also cover literal usages for the given set of idiomatic expressions. We also present the utility of our dataset by using it to train a sequence labelling module and testing on three independent datasets with high accuracy, precision and recall scores.

\end{abstract}
\section{Introduction}
Natural language understanding of idiomatic expressions embedded in sentences has been a complex problem to solve for some time. Idiom handling has been a problematic area for a variety of NLP tasks. \cite{volk1998automatic}, \cite{sag2002multiword} and \cite{cap2015account} have discussed the magnified complexity of this problem with respect to linguistic precision. \cite{salton2014empirical} provides empirical evidence that state-of-the-art machine translation systems may achieve only half of the BLEU score on sentences that contain idiomatic expressions as compared to the ones that do not. This drop in the score occurs not only due to the comparatively low frequency of the idiomatic phrase with respect to the frequency of the constituent words, but also due to the lack of automatically determinable clear patterns in the wide and varied instances of idioms in data \cite{fillmore1988regularity}. This makes a regular monolingual training dataset sparse with respect to idiomatic expressions. The absence of a dataset rich in idiomatic expressions hampers the possibility of modelling the problem into a machine learning task. 
\newline Any attempt on handling these idiomatic expressions has to follow certain predefined steps as discussed in \cite{liu2016phrasal}. The first step is to detect lexical occurrences of idiomatic expressions in a given text. The subsequent steps constitute identifying the underlying semantics and learning a simpler representation for any downstream task. In this paper, we attempt the first step from the aforementioned steps i.e. detection of possible idiomatic expressions in a given text. These lexical variations can have a literal occurrence as our purpose is to capture the span of the phrase in order to identify a metaphorical usage as the next step. We present a dataset of 25206 sentences which contain lexical occurrences of 717 idiomatic expressions from the IMIL dataset \cite{agrawal2018no}. We identify the detection of idiomatic expressions as a sequence labelling task and present a two pronged approach for detection of two different kinds of idioms: Static and Formal. Static idioms do not undergo lexical changes, therefore labelling them can be as simple as a string search in the text. Formal idioms, on the other hand, undergo various lexical modifications, therefore labelling them can be modelled as a supervised task. We test a model trained on our dataset and test on three datasets,"all words" and "lex sample" training datasets of SemEval-2013 Task 5b Dataset\cite{korkontzelos2013semeval}, and PIE Corpus\cite{haagsma2019casting}. All tests give results with high accuracy, precision and recall scores.

The major contributions of this work can be summarized as follows:

\begin{itemize}
    \item We publically release a dataset of 25206 sentences labelled with lexical occurrences of 717 idioms. These labels are done by automatic systems with high accuracy. Of these, 21891 sentences contain occurrences of Static idioms which are 359 in number and 3135 sentences contain occurrences of Formal idioms which are 358 in number.\footnote{Dataset available at: https://github.com/prateeksaxena2809/EPIE\_Corpus}
    \item An analysis of the distribution(Mean and Standard Deviation) of idioms over the dataset.
\end{itemize}

\section{Related Work}

\cite{fillmore1988regularity} created a distinction in idioms i.e. Formal and Static. Static idioms are the kind of idioms which do not exhibit internal or morphosyntactic variation. For example, \textit{As soon as possible}, \textit{no comment}, etc. Formal idioms, on the other hand, undergo inflectional changes, pronominal and determiner modifications, and internal qualitative modifiers(adjectival and adverbial). For example, \textit{keep eye on}, \textit{race against time} etc.
StringNet\cite{wible2010stringnet} identified that mapping base forms of phrases is necessary in order to extract their surface realization. StringNet used hybrid ngrams and cross indexing to create a resource to extract idiomatic sentences from the British National Corpus corpus\cite{leech1992100}. We use StringNet for the first level extraction of sentences for our work.
\cite{agrawal2018no} has created the IMIL dataset which maps 2000 of the highly occurring English idioms to their counterparts in different Indian languages. We use their idiom list as a starting point for our sentence extraction.

There have been some attempts to extract idiomatic expressions. The VNC-Tokens Dataset\cite{cook2008vnc}, IDIX Corpus\cite{sporleder2010idioms}, PIE Corpus\cite{haagsma2019casting} and SemEval-2013 Task 5 Dataset\cite{korkontzelos2013semeval} all contain around 3000 to 4500 potential idiomatic expressions instances of 53 to 65 candidate idioms. These datasets, though thorough for their respective candidate idioms, are small in size and limited in coverage. Our dataset attempts to provide a wider coverage over a larger dataset.   

\section{Data}

Our aim is to create a dataset only containing sentences with lexical occurrences of idioms for the IMIL dataset. This requires multiple data filtering steps. These steps are explained in the subsequent subsections.

\subsection{StringNet Extraction}
Variations in Idiomatic Expressions occurs in the following forms:
\begin{itemize}
    \item Inflectional Modifications (tense, gender, number, etc):
    
    \emph{Bite the dust}
    \begin{itemize}
        \item  The visiting team \emph{bit the dust} in the football game yesterday.
    \end{itemize}
    \item Determiner/Pronominal Replacement:
    
     \emph{Keep up the good work}
    \begin{itemize}
        \item \emph{Keep up your good work} and the promotion will follow.
    \end{itemize}
    \item Named Entities and Qualitative Modifiers inclusions(Adjectival and Adverbial)
    
        \emph{Keep an eye on}
    \begin{itemize}
        \item \emph{Keep a keen eye on} the child while he plays.
    \end{itemize}
        \emph{Behind his back}
    \begin{itemize}
        \item People say a lot \emph{behind James' back}.
    \end{itemize}
\end{itemize}
In order to extract all instances of an idiomatic expression, it is important to account for all the variation in the expression. We use StringNet for this task. Stringnet contains two billion connected hybrid ngrams cross-indexed with lexeme information, parts of speech information and various word forms. This matches an idiomatic expression like \textit{keep your eye on} to its inflectional modifications like \textit{kept your eye on} and \textit{keeps your eye on}. We also utilize StringNet's unique feature of vertical pruning and horizontal pruning. Vertical pruning refers to generalization of lexemes in a given search entry in order to search occurrence of parent ngrams and child ngrams of the entry in the corpus. For example, a parent ngram of the entry \textit{Keep your eye on} is \textit{keep [pron] eye on} as [pron] constitutes all pronouns. Vertical Pruning helps in extraction of pronominal and determiner variation. Horizontal pruning refers to connecting an ngram with another ngram which differs by one unit or type of ngram. For example, the entry \textit{keep [det] eye on} can be connected to \textit{keep eye on} and \textit{keep [det] keen eye on} using horizontal pruning because it differs from these ngrams by a length of 1. But the entry \textit{keep your eye on} can also be connected to \textit{keep an eye on} using horizontal pruning because both entries differ by 1 ngram type. Horizontal pruning helps in extraction of determiner-pronoun interchangeability and internal qualitative modifiers.

We take the 2000 idioms present in the IMIL dataset and process them automatically in order to be used as search entries into StringNet. The processing involves two features; lemmatization, and generalization of pronouns and determiners into generic entries \textit{[pron]} and \textit{[det]} respectively. An entry \textit{keep an eye on} becomes \textit{keep [det] eye on}. In addition to searching the term, we also search the idiom in both directions through one level each of vertical and horizontal pruning. This results in the extraction of 81562 sentences containing instances from 758 of the 2000 idioms.

\subsection{Candidate Idioms Selection} 
In this step, we filter out redundant idioms from our idioms list
Redundant idioms constitute similar idiom entries in the 758 idioms list like \textit{music to my ears} and \textit{music to my ear} are clubbed into a single entry, removing duplicate entries of instances from the sentences. This step results in filtering 749 idioms and 77894 sentences. The idioms that remain are unique and have idiomatic usages.

\subsection{Candidate Instances Selection}
Idiomatic Expressions are also idiosyncratic in the kind of lexical variations they allow. In this step, we filter out those lexical variations of idioms, which will never occur idiomatically. This requires extraction of specific patterns which are relevant exclusively to particular idioms. For example, the idiom \textit{keep an eye on} can occur as \textit{keep your eye on} but \textit{give me a hand} cannot occur as \textit{give me your hand}. In order to efficiently extract correct patterns, we manually divide the idioms list into two categories based on  \cite{fillmore1988regularity}.
\subsubsection{Static Idioms} 
Static idioms are idioms which do not undergo any lexical modification. We identify 388 idioms as Static in our idioms list. These idioms have 45955 instances in the data. We filter out sentences which did not have an exact occurrence of the idiom. If no exact occurrence of an idiom is found, we reject the idiom altogether. At the end of this step, 21891 sentences with 359 Static idioms are left.
\subsubsection{Formal Idioms}
Formal Idioms are idioms which occur in sentences with various lexical modifications. We identify 361 idioms from our idioms list as Formal idioms based on their occurrences. These idioms have 31939 instances in the data. As this task requires more flexibility and complexity than Static idioms, an completely automatic approach is not feasible. At the same time, going through the whole dataset sentence by sentence is quite inefficient. Thus, in order to efficiently sift through the data, we extract the unique variations of each idiom and then manually remove the irrelevant occurrence patterns, thus removing all sentences with those occurrences. This reduces our load by a scale factor of 1/3 as the unique occurrences are around 10000 in number. This process does not reduce the number of idioms to large extent(358) but we do filter out a considerable number of patterns, resulting in only 3135 remaining sentences.

\subsection{Final Result}
Finally we create a dataset of 717 idioms in 25026 sentences/instances. We separate the data into two groups; Static and Formal idioms. We create this distinction in our data because detection of both categories of idioms require separate steps. Static idioms can be detected by treating them like words-with-spaces and simply finding their exact matches in the sentence. Formal idioms detection requires a more complex approach which can identify the similarities between instances of the same idiom and their difference from other phrases. Number of sentences and idioms left after each step are given in Table \ref{tab:data}. The first three rows show the results for the total data extraction while the subsequent rows show extraction results for Formal and Static idioms separately.

\begin{table*}[t!]
\centering
\begin{tabular}{|c|c|c|}

      \hline
      \textbf{Extraction Step}&\textbf{Sentences}&\textbf{Idioms} \\
      \hline
      StringNet Extraction & 81562 & 758\\
      \hline
      Candidate Idioms Selection(Total)  & 77894 & 749\\
      \hline
      \textbf{Candidate Instances Selection(Total)} & \textbf{25206} & \textbf{717}\\
      \hline
      Candidate Idioms Selection(Static Idioms)  & 45955 & 388\\
      \hline
      \textbf{Candidate Instances Selection(Static Idioms)} & \textbf{21891} & \textbf{359}\\
      \hline
      Candidate Idioms Selection(Formal Idioms)  & 31939 & 361\\
      \hline
      \textbf{Candidate Instances Selection(Formal Idioms)} & \textbf{3135} & \textbf{358}\\
      \hline
\end{tabular}
\caption{Number of Sentences and Idioms left after each extraction step}
\label{tab:data}
\end{table*}

\begin{table*}[t!]
\centering
\begin{tabular}{|c|c|c|c|}

      \hline
      \textbf{Test Dataset}&\textbf{Accuracy}&\textbf{Precision} &\textbf{Recall}\\
      \hline
     Formal Idioms Test Dataset & \textbf{0.98} & \textbf{0.95} & \textbf{0.91}\\ 
      \hline
     SemEval All Words Dataset(all usages) & 0.84 & 0.90 & 0.85\\
     \hline
     SemEval All Words Dataset(idiomatic usages) & 0.86 & 0.93 & 0.86\\
     \hline
     SemEval Lex Sample Dataset(all usages) & 0.89 & 0.90 & 0.90\\
     \hline
     SemEval Lex Sample Dataset(idiomatic usages) & 0.92 & 0.95 & 0.92\\
     \hline
     PIE Corpus(all usages) & 0.69 & 0.60 & 0.69\\
     \hline
     PIE Corpus(idiomatic usages) & 0.88 & 0.94 & 0.88\\
     \hline

\end{tabular}
\caption{Test Results from the model trained on Formal Idioms Training Dataset. Formal Idioms Test Dataset is 25\% split from the Formal Idioms Dataset. All datasets have been tested separately for \emph{All Usages} and \emph{Only Idiomatic usages} of potentially idiomatic expressions in sentences}
\label{tab:results}
\end{table*}

\begin{table}[t!]
\centering
\begin{tabular}{|c|c|c|c|}

      \hline
      \textbf{Idiom Type}&\textbf{Sentences}&\textbf{Mean}&\textbf{Std Dev} \\
      \hline
      Formal & 3135 & 8.75 & 8.61\\
      \hline
      Static & 21891 & 60.9 & 160\\
      \hline
\end{tabular}
\caption{Mean and Standard Deviations of Final Datasets}
\label{tab:meanstd}
\end{table}

We are also interested in finding the spread of each idiom in our idioms list. In this effort, we calculate the total instances of each idiom and calculate the mean and standard deviation on the resultant counts respectively for Formal idioms and Static idioms. Table \ref{tab:meanstd} shows the mean and standard deviation of both the Formal idioms dataset and Static idioms dataset with respect to their number of occurrences in data. The mean and standard deviation for Formal idioms are very close which suggests an exponential distribution whereas the Static idioms show a skewed distribution.

\section{Experiments}
We use our Formal idioms dataset containing 3135 sentences to train on a typical sequence labelling neural network.
We do a 75-25 train-eval split on our dataset for our training and evaluation. In addition to the Formal idioms test dataset, we use three independent datasets for testing mentioned as follows:
\begin{itemize}
    \item "All words" training dataset from \cite{korkontzelos2013semeval} containing 1143 sentences. All sentences contain potentially idiomatic phrases, each usage is labelled with \emph{idiomatic},\emph{literal} or \emph{both} usage.
    \item "Lex sample" training dataset from \cite{korkontzelos2013semeval} containing 1423 sentences. All sentences contain potentially idiomatic phrases, each usage is labelled with \emph{idiomatic},\emph{literal} or \emph{both} usage.
    \item PIE corpus\cite{haagsma2019casting} containing 2239 sentences. All sentences contain potentially idiomatic phrases, each usage labelled with a sense label,"y" meaning idiomatic usage and "n" meaning literal usage.
\end{itemize}
We evaluate our models on two versions of each of the three datasets: All samples and samples labelled with idiomatic usages. 
 
We use a BiLSTM-CRF \cite{DBLP:journals/corr/HuangXY15} module for our task. We use 300 dimensional glove embeddings\cite{pennington2014glove} as our embedding input. We use LSTM hidden representation of dimension 100 and batch size of 20. We train the model for 25 epochs.

\section{Results}
The Results can be seen in Table \ref{tab:results}. We see that the Formal idioms test dataset gives the best results because of similarity with the training dataset. However, the model also gives good results with other independent datasets.

\section{Conclusion}
In this paper, we present a semi-automatic approach to create a new dataset of labelled potentially idiomatic expressions in 25206 English Sentences extracted from the BNC corpus\cite{leech1992100} with high accuracy. We segregate our dataset into two categories, Formal and Static. This we do because of the difference in the potentially idiomatic span detection mechanisms of these categories.

%
%
%
%
%
%
\bibliography{tsd048}
\bibliographystyle{tsd048}




\end{document}